\documentclass[conference]{IEEEtran}
\IEEEoverridecommandlockouts
\usepackage{cite}
\usepackage{amsmath,amssymb,amsfonts}
\usepackage{algorithmic}
\usepackage{graphicx}
\usepackage{textcomp}
\usepackage{xcolor}
\usepackage{subfig}
\def\BibTeX{{\rm B\kern-.05em{\sc i\kern-.025em b}\kern-.08em
    T\kern-.1667em\lower.7ex\hbox{E}\kern-.125emX}}

\usepackage{color}
\usepackage[colorinlistoftodos]{todonotes}
\usepackage{soul,xcolor}

\usepackage[hyphens]{url}
\usepackage[hidelinks]{hyperref}
\hypersetup{breaklinks=true}

\begin{document}
\newcommand{\ahsan}[1]{[{\color{red}Ahsan: #1}]}
\newcommand{\naresh}[1]{[{\color{blue}Naresh: #1}]}
\newcommand{\kecheng}[1]{[{\color{brown}Kecheng: #1}]}
\title{Dynamic Heuristic Neuromorphic Solver for the Edge User Allocation Problem with Bayesian Confidence Propagation Neural Network\\
\thanks{This work is funded by the European Union under the call HORIZON-CL4-2023-HUMAN-01-01 Research \& Innovation Action (RIA), Grant no. 101135809 (EXTRA-BRAIN).}
}

\author{ Kecheng Zhang$^{1}$, Anders Lansner$^{1,5}$, Ahsan Javed Awan$^{2}$, Naresh Balaji Ravichandran$^{1}$, Pawel Herman$^{1,3,4}$ \\ \small $^1$KTH Royal Institute of Technology, $^2$Ericsson Research, $^3$Digital Futures, $^4$Swedish e-Science Research Centre, $^5$Stockholm University \\ kechengz@kth.se, ala@kth.se, ahsan.javed.awan@ericsson.com, nbrav@kth.se, paherman@kth.se }
\maketitle
\begin{abstract}
We propose a neuromorphic solver for the NP-hard Edge User Allocation problem using an attractor network with Winner-Takes-All (WTA) mechanism implemented with the Bayesian Confidence Propagation Neural Network (BCPNN) framework. Unlike previous energy-based attractor networks, our solver uses dynamic heuristic biasing to guide allocations in real time and introduces a ``no allocation" state to each WTA motif, achieving near-optimal performance with an empirically upper-bounded number of time steps. The approach is compatible with neuromorphic architectures and may offer improvements in energy efficiency.


\end{abstract}
\section{Background}

Edge computing enables low latency and energy-aware services by placing computation closer to users\cite{informatics11040071},\cite{cao2024cost}. This shift introduces the Edge User Allocation (EUA) problem, which involves selecting user-to-server assignments under strict capacity and coverage constraints. Exact optimization methods quickly become computationally expensive as problem size grows, while approximate solvers running on conventional architectures still incur substantial energy costs\cite{qubo}. Neural models often require size-specific training \cite{bello2016neural} while struggling to encode EUA's discrete and multi-resource constraints. Neuromorphic approaches present a promising alternative for energy-constrained environments\cite{SNNhardwareReview},\cite{KARAMIMANESH2025107256}, but existing models face limitations in scalability and often ignore the possibility of leaving users unassigned, when edge server resources are insufficient to handle all requested tasks within its coverage area, which can lead to infeasible solutions. In this work, we propose a solver that is based on the BCPNN framework. The network is organized as a collection of modular WTA circuits, with one WTA module per user, and is driven by dynamic heuristic input. Our model explicitly includes ``no allocation" units, updates bias signals over time to guide search, and scales to larger instances without the need for parameter tuning. The structure is designed to be compatible with future neuromorphic deployment.

\subsection{The Edge User Allocation Problem}


Emerging edge–cloud applications, such as collaborative intelligence, autonomous driving, and augmented/virtual reality, offload computation-intensive tasks from user devices to edge servers, keeping devices lightweight and energy-efficient. Edge resources are typically distributed across communication service provider (CSP) infrastructure~\cite{ericsson_edge} and may extend to radio base stations to support ultra-low-latency services. 

User devices requests to offload computation tasks with specific resource requirements that can be executed on any suitable edge server. At the same time, each device may be within the coverage of multiple edge servers. The main challenge is to determine which device should be assigned to which edge server in order to maximize the number of served UEs, while ensuring that the total resource demand of allocated UEs does not exceed the capacity of the chosen server. This challenge is referred to as the Edge User Allocation (EUA) problem. 
In practical deployments, assignments may need to be dynamically adjusted to account for changes in UE locations, variations in the number of offloading requests, fluctuations in individual UE resource demands, and changes in available edge server resources. In this work, however, we consider a static snapshot of the system, where UE locations, resource demands, and edge server capacities are assumed to remain fixed during the allocation process.


The EUA problem can be formulated as a multi-objective binary combinatorial optimization problem that involves $n_u$ users and $n_s$ servers. Due to the nature of our model framework, the variables to optimize are $x_{i j} $, which are binary variables indicating whether user $i$ is allocated to server $j$. In particular, $x_{i (n_s+1)}$ means that the user $i$ is not allocated to any server.
\begin{figure}[htbp]
\centerline{\includegraphics[width=\linewidth]{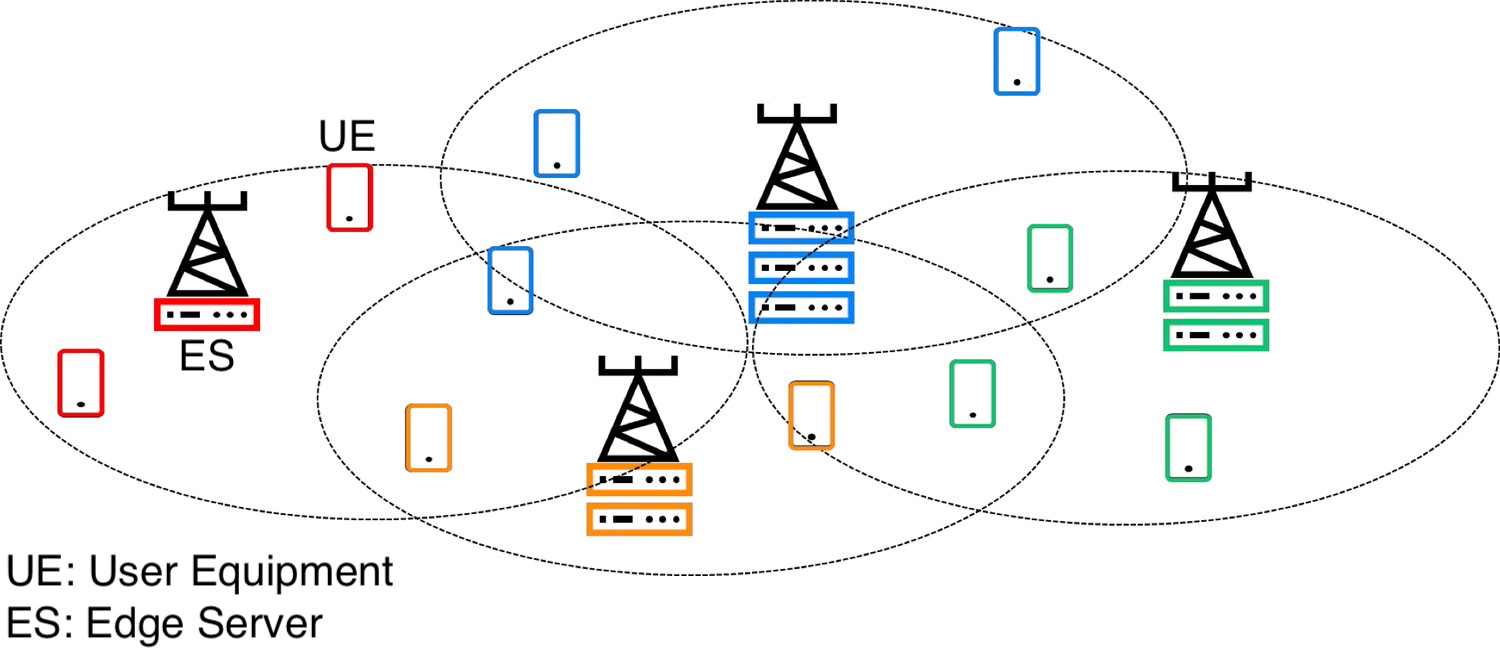}}
\caption{A geographical visualization of the Edge User Allocation Problem. The figure shows multiple Edge Servers (ESs), and a set of User Devices (UEs) distributed across a geographical area. Each base station has a limited coverage range (dotted ellipses), within which it can potentially serve users. Different colors represent separate edge servers and their assigned users. The core challenge is to determine which users should be assigned to which servers, under capacity and coverage constraints.} 
\label{fig}
\end{figure}

The optimization objectives are
\begin{equation}
\begin{aligned}
\text{maximize: } A_u &= \sum_{i = 1}^{n_u} \sum_{j = 1}^{n_s} x_{ij},\\
\text{minimize: } U_s &= \sum_{i = 1}^{n_u} \operatorname{sgn}\!\left( \sum_{j = 1}^{n_s} x_{ij} \right)
\end{aligned}
\end{equation}
Subject to
\begin{equation}
\sum_{i = 1}^{n_u} d_i x_{ij} \le c_j, \quad \forall j = 1, \ldots, n_s
\end{equation}
\begin{equation}
\sum_{j = 1}^{n_s + 1} x_{ij} = 1, \quad \forall i = 1, \ldots, n_u
\end{equation}
\begin{equation}
x_{ij} \le \mathrm{cov}(u_i, s_j)
\end{equation}
Where $A_u$ stands for the number of allocated users, and $U_s$ means the number of servers that are used. 
$n_u$ and $n_s$ are the number of users and servers in the current problem case.
$d_i$ is the resource demands of user $i$. Note that this is a vector containing two elements, each for a different resource. Similarly, $c_j$ is the capacity vector of server $j$. For the data used in this work we have $d_i = [d_i^\text{core}, d_i^\text{ram}]$ and $c_j = [c_j^\text{core}, c_j^\text{ram}]$.
$\text{cov}(u_i , s_j)$ is a function that indicates whether user $i$ is within the coverage of server $j$.

Solving the EUA problem optimally is NP-hard because, in the simplest setting with a single resource dimension and identical bins, EUA reduces to the classical bin packing problem, which is known to be NP-hard~\cite{EUAbinpacking}. It quickly becomes intractable with conventional mixed-integer programming solvers. Many approximation methods have been proposed:
Phu Lai et al. \cite{EUAbinpacking} modeled EUA as a variable-sized vector bin packing problem and solved it via lexicographic goal programming. He et al. \cite{EUAgame} treated EUA as a potential game where each user selects a server to maximize utility, achieving a distributed Nash equilibrium. Li et al. \cite{EUAFOA} applied a Fruit Fly Optimization Algorithm (FOA) to search for balanced user–server allocations. Extensions to further relax resource constraints or enable partial allocations have been proposed \cite{EUAQoS}.
While these approaches improve efficiency compared to greedy baselines, they still rely on conventional von Neumann architectures, which limit scalability and energy efficiency in edge environments.

Applying machine learning to EUA remains challenging. Tarasova \cite{Anna} formulated EUA as a seq2seq problem using LSTM and Transformer models, but the outputs required post-processing to satisfy constraints. More general neural combinatorial optimization frameworks, such as pointer networks \cite{bello2016neural}, Graph Neural Networks \cite{GNN}, Message Passing Neural Networks \cite{MPNN}, and attention-based models \cite{attention}, have been investigated. However, they struggle to encode EUA’s discrete coverage and multi-resource capacity constraints, and models often need retraining for each problem size.

Another recent yet promising approach is the neuromorphic method due to its energy efficiency. The first attempt on the EUA problem is proposed by Petersson Steenari \cite{PeterssonSteenari1701277}, in which each user's all possible allocations are modeled together as a winner-takes-all motif. While the architecture naturally fits the EUA setting, this type of model requires careful parameter tuning to achieve good performance. Therefore, most of the subsequent works focus on designing structural or algorithmic ways to achieve automatic parameter tuning while reusing or slightly extending that basic circuit. Falkeström \cite{Falkestrom1770572} leaves the circuitry unchanged and introduces a Bayesian Artificial Bee Colony (ABC-BA) optimizer that searches the network parameters. Allard \cite{Allard1837710} adds three specialized neurons—Control, Deterministic-Capacity, and Monitor that hardwire the constraints, shrink the search space, and supply an internal fitness signal. A self-adaptive genetic algorithm (GA-SSNN) then evolves the remaining parameters. Schmützer \cite{Gustav} models EUA as a QUBO, and all problem constraints are embedded into a quadratic energy function that also determines the weight matrix of the model. Monitor neurons are introduced to guide the parameter update in a parallel simulated annealing style, which raises the issue that if two or more variables fulfill the Boltzmann condition, their joint flipping can worsen the solution due to interaction terms. This could lead to sub-optimal solution quality to the problem expressed in QUBO form compared to the state of the Gurobi solver solving the same problem expressed in ILP form \cite{qubo}.

This line of neuromorphic works shares two major limitations. The first is scalability; all the forementioned works reported an obvious decrease in performance when handling larger problem instances because they require computationally expensive parameter tuning schemes based on global history of past evaluations. The second problem is that they only consider user-server pairs as a neuron, while ignoring that, in certain cases, not all users can be allocated. Assigning each user to a server in such cases will lead to invalid or low-quality solutions.


To address these two problems, firstly, we introduced a "no allocation" unit into each WTA motif, which allows the network with insufficient server resources to serve the requests from a subset of users at once. And then we design a set of heuristic functions that generate the external input for all neurons that changes dynamically in each simulation timestep.

\subsection{BCPNN as WTA motifs}

We implement the WTA motifs with a simple instance of Bayesian Confidence Propagation Neural Network (BCPNN), which is a cortex-inspired ANN framework \cite{BCPNNorigin},\cite{BCPNNrepspike}. Its neural activations encode the likelihood of input features or categories being present, while synaptic connections reflect estimated correlations between units. The resulting activity dynamics effectively approximate the computation of posterior probabilities. Under the BCPNN framework, the update process of each WTA motif is introduced below.

In timestep $t$, every unit (or minicolumn) receives input from all connected units' activation, and also an external input $I$ that can be from the training data and manually set.
\begin{equation}
s_i'^{(t)} = b_{i j} + I_i + \sum_j^M w_{i j} \cdot a_j^{(t - 1)}
\end{equation}
Then the unit support value is obtained by taking an exponential moving average.
\begin{equation}
s_i^{(t)} =(1 - \alpha) \cdot s_i^{(t - 1)} + \alpha \cdot s_i'^{(t)}
\end{equation}
Then a softmax function is applied to get the unit confidence, which is the probability of this feature being true.
\begin{equation}
\pi_i = \frac{e^{s_i}}{\sum_j^M e^{s_j}}
\end{equation}

Then within each hypercolumn (WTA motif), an activation function $f$ is applied, e.g., WTA, Stochastic WTA.
\begin{equation}
a_i^{(t)} = f(s_i^{(t)})
\end{equation}


A Stochastic Winner-Take-All (WTA) mechanism is used during the inference stage. 
At each timestep, exactly one unit within each hypercolumn is activated, sampled according to the confidence values $\pi_i$. 
This results in a one-hot activation vector $\mathbf{a}^{(t)}$ drawn from a categorical distribution for each hypercolumn:

\begin{equation}
\mathbf{a}^{(t)} \sim \text{Categorical}(\boldsymbol{\pi}^{(t)}).
\end{equation}

\section{Methods}

Our proposed model consists of a BCPNN and a dynamic heuristic generator that interact with each other. The potential allocation states of each user is modeled as a WTA motif that consists of $n_s+1$ units, where $n_s$ corresponds to the number of servers and the additional unit represents the option for the user to remain unallocated. The network operates over discrete simulation timesteps. At each step, the current activation pattern is interpreted as a candidate user-to-server assignment. Based on this assignment, server utilization statistics are computed and passed to a dynamic heuristic generator, which in turn produces external input signals that modulate the activation dynamics of each unit in the next timestep. The model iteratively evolves until termination criteria are met. 


\begin{figure}[htbp]
\centerline{\includegraphics[width=\linewidth]{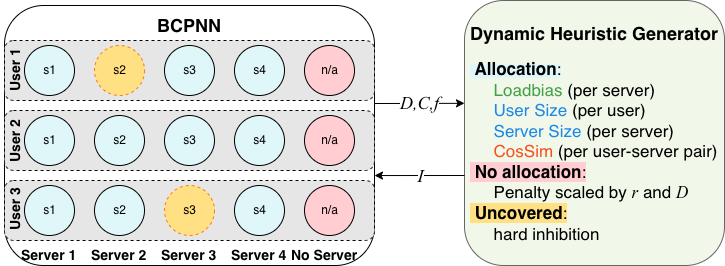}}
\caption{Overview of the proposed BCPNN--EUA model. 
Each row in the left panel represents a user and forms a WTA hypercolumn whose units correspond to the possible allocation states: being assigned to one of the servers (s1--s4) or left unallocated ("n/a"). 
Blue circles denote allocation units, the pink "n/a" units implement the explicit no-allocation option. 
yellow circle indicates that the user is outside the coverage of the server.
The right panel shows the dynamic heuristic generator, which computes external input $I$ for all units based on relative demand, relative capacity, and server fill degrees $(D, C, f)$. }

\label{fig}
\end{figure}


The external input module adopts a dynamic heuristic approach and applies different procedures for different types of units. The units are categorized into three types: allocation units, no allocation units, and units outside the coverage of all servers (uncovered units).

\subsection{The Dynamic Heuristic Generator}
We categorize the units by: uncovered units, allocation units, and no-allocation units, and we design the heuristic input functions for them separately.

To guide the allocation dynamics, we design external inputs based on numerical features that reflect resource pressure and server utilization. Two key features are:

\begin{itemize}
    \item Demand-Capacity Ratio (DC ratio): Measures overall resource scarcity by comparing total user demand to total server capacity:
    \begin{equation}
    \boldsymbol{r} = \frac{\sum_{u=1}^{n_u} \boldsymbol{d}_u}{\sum_{s=1}^{n_s} \boldsymbol{c}_s}, \quad \boldsymbol{r} = [r^{\text{core}},\ r^{\text{ram}}]
    \end{equation}

    \item Server Filled Degree: Indicates the utilization level of each server:
    \begin{equation}
    \boldsymbol{f}_j = \frac{\sum_{i=1}^{n_u} x_{ij} \cdot \boldsymbol{d}_i}{\boldsymbol{c}_j},\quad \boldsymbol{f} = [f^{\text{core}},\ f^{\text{ram}}]
    \end{equation}
\end{itemize}

These features are recomputed at every timestep and used to dynamically compute and deliver the external input to each allocation unit.

\subsubsection{Input Design for uncovered units}

For an allocation unit $x_{ij}$ whose user $i$ is not covered by its server $j$, it should never become activated since that violates the coverage constraint. To achieve so, we assign a large negative value to all uncovered units, so they will have a near-zero probability of getting activated. In practice, we set the value as
\begin{equation}
I_{i j} = 2 * y_{\min}
\end{equation}
Muting illegal assignment via setting input is an advancement compared to previous models since it provides an elegant way of excluding the illegal states from the search space, in contrast to including them in the search space but assigning them huge penalties. 

\subsubsection{Input Design for Allocation Units}

The first component is the Loadbias Curve. An obvious intuition of a good EUA solution is: If a server is open, it should be used as fully as possible. The loadbias function formalizes this idea by assigning a higher value when a server is closer to being almost fully filled and a lower value for server overfilling and underutilization. 

\begin{figure}[htbp]
\centerline{\includegraphics[width=\linewidth]{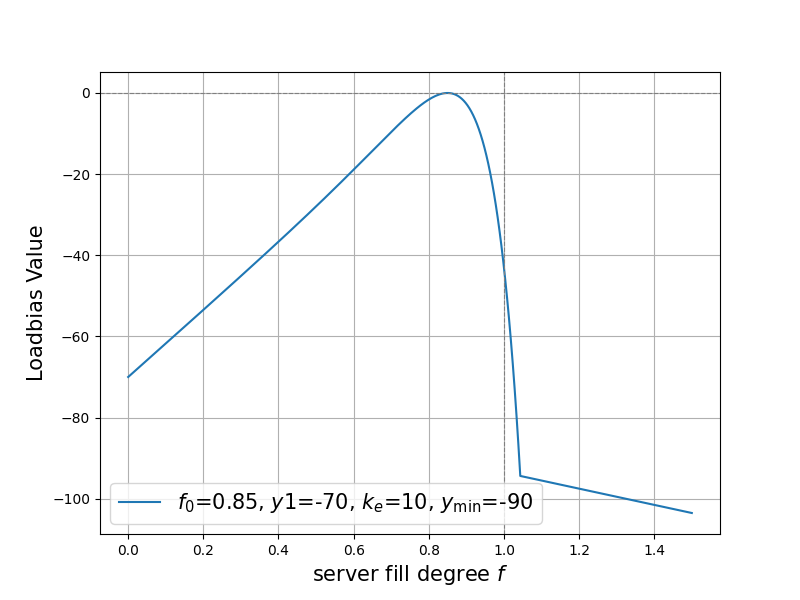}}
\caption{The loadbias curve} 
\label{lbcurve}
\end{figure}

To capture this intuition, we design a loadbias function as shown in figure \ref{lbcurve}. It is defined as below. 

First, we start with some helper functions:

\begin{equation}
y_\mathrm{exp}(f)  = y_1 \frac{1 - f}{f_0} \cdot (1 - e^{k_e f})
\end{equation}

\begin{equation}
f_c \colon y_{\mathrm{exp}}(f_c) = y_{\min}
\end{equation}

\begin{equation}
y_\mathrm{lin}(f) = y_{\min} + k(f - f_c)
\end{equation}

And the loadbias function is defined as
\begin{equation}
\text{loadbias}(f) = \begin{cases}
y_\text{exp}(f) \text{, } f \le f_c \\ y_\text{lin}(f) \text{, } f > f_c
\end{cases}
\end{equation}
The function has five tunable and explainable parameters:
\begin{itemize}
    \item $y_1$: The value assigned to 0 server usage
    \item $y_{\min}$: The lowest value to take from $y_\text{exp}$
    \item $f_0$: The x-coordinate for the peak of the curve, or the most encouraged server fill degree.
    \item $k_e$: The "steepness" of the curve
    \item $k$: The slope of $f_\text{lin}$ For the calculation of the loadbias value, $f = \max(f^\text{core}, f^\text{ram})$.
\end{itemize}

The second component considers the size of users and servers. A desirable allocation should give priority to smaller users when the system is short of resources (that is, a high $r$) and if not to use all servers, open the larger ones first, since both strategies help maximize $A_u$ with a given $U_s$ and a fixed server fill degree.

The ``largeness" of a user or server needs to be defined since demands and capacities are 2D-vectors, while the input of a unit must be a scalar. Here we make a simplification and just use the number of CORE to describe the size of a user/server, because for almost all the test cases in the dataset, CORE is more in demand than RAM. However, without loss of generality, we can define largeness as the resource that has the highest demand-capacity ratio. Another potential option would be a summed weight of both resources with weight proportional to their dc-ratio, although it would take test cases with more variety to examine the utility of these definitions.

The relative user demand and relative server capacity are used to describe how large a user/server is compared with the average user/server size of the case. They are defined as
\begin{equation}
D_i = \frac{n_u \cdot d_i}{\sum_k^{n_u} d_k} - 1
\end{equation}

\begin{equation}
C_j = \frac{n_s \cdot c_j}{\sum_k^{n_s} c_k} - 1
\end{equation}

This definition gives $D$ and $C$ an average of zero while describing their relative sizes. Zero average is a useful property here because it allows the importance of the component to be varied without changing the average of the full input or the relative value between the supports of allocation and no allocation units.

Therefore, $D^\text{core}$ and $C^\text{core}$ are used as components of the input.

Even though the size of entities is defined by a single resource, it does not mean the other resources can be simply ignored. It is observed that when both resources are in high demand and only one of them, in this case, CORE, is considered, the servers may get filled up with RAM before the CORE capacity is reached. 

The capacity constraint needs to be satisfied for both resources simultaneously. Previous work approached this dual-constraints problem by encoding both as separate terms in the energy function. This, however,  collapses the two-dimensional constraint into a static, weighted sum and thus it lacks flexibility.
To ease this problem, we propose a dynamic method inspired by bin packing problem \cite{tetris},\cite{Binheuristics}, which makes dynamic, timestep-by-timestep adjustments to the unit input in order to saturate both capacities as much as possible.

In each timestep, we calculate the cosine similarity $theta$ of each user demand vector $d$ with each server remaining space vector, formally
\begin{equation}
\theta_{i j} = \frac{d_i \cdot (c_j - \sum_k^{n_u} x_{k j})}{|d_i | \cdot |c_j - \sum_k^{n_u} x_{k j} |} 
\end{equation}

Combining equations from this section, we have the input for an allocation unit as
\begin{equation}
I_{i j} = \text{loadbias}(f_j) - k_1D_i^\text{core} + k_2C_j^\text{core} + k_3 \theta_{i j} - b_3
\end{equation}
Where $k_1, k_2, k_3$ are tunable parameters, representing the importance of the corresponding component. Note that $D$ and $C$ have an average of zero. It is hard to predict the average of all cosine similarity terms. Therefore, we estimate it by the medium of the range of its value $[0,1]$, and then shift the medium to zero, by setting  $b_3$ to be half of $k_3$. 

\subsubsection{Input Design for no-allocation units}

The support value of the no-allocation units is to set a benchmark of whether it is "worth it" to allocate this user. Two factors are considered here: On a global level, the scarcer the server resources are, the more likely we are to lose more users. At an individual level, the more demanding a user is, the more likely we are to give up serving them when there is a lack of resources.

Following these principles, we formalize as
\begin{equation}
I_{i (n_s + 1)} = k_0r^{\text{core}} D_i + b_0
\end{equation}
Where $k_0$ and $b_0$ are tunable parameters. Note that $b_0$ plays a crucial role in the model, as it controls the baseline for retaining a user, thereby roughly determining the tradeoff between servers and users, or the percentage of servers that should be used.

\subsection{Model Execution}
We set the self-connection weight of each unit to a positive constant, and within each WTA motif, every two different units share a negative constant connection weight.

The model needs to be run multiple times with different values of $b_0$, in order to fully explore the different tradeoff levels between users and servers. $b_0$ should be selected from a set of predetermined values, between

\begin{itemize}
    \item $b_0=0$, in this situation, the no-allocation unit will almost always get activated for all user hypercolumns, resulting in 0 server usage and 0 user allocation, or a ``server-greedy" solution;
    \item $b_0$ is sufficiently low, e.g., $b_0=2y_{min}$, in this case, the no-allocation units will rarely get activated unless all servers are full, leading to a ``user-greedy" solution.
\end{itemize}

The network is reset for every different value of $b_0$, the current best solution found is maintained and updated for each $b_0$. An obvious limitation of this design is that it multiplies the timesteps needed to find a solution, but in our experiments the solver still converges to good solutions within a few hundred timesteps per instance.

For the termination criteria of each $b_0$, the network stops updating when the network activation remains for 5 continuous timesteps or a threshold number of timesteps is reached.

After the network stops updating, the activation matrix of the final timestep is read out and translated into an allocation solution. Since the loadbias curve only imposes soft constraints on server capacity, servers may get slightly overfilled. If a server is overfilled, the largest user assigned to it would be taken out, and the process repeats until there are no overfilled servers, resulting in the final solution.

\subsection{Evaluation Dataset and Experimental Settings}
The dataset used for evaluation of performance in this project was synthesized using the EUA data generator proposed by Tarasova\cite{Anna}. The dataset consists of 30 cases. It includes the coordinates of users and servers, the resource demand of each user, the capacity of each server, the coverage range of each server, and the set of covered users. Two types of resources are considered in the dataset, which are the number of CPU cores and the amount of RAM. 
There are two types of EUA cases in the dataset, namely centralized and distributed ones. A distributed case corresponds to a standard case defined above, while the centralized cases represent the real-world setting where a proxy is used to receive the requests of the users and forward them to a group of centralized servers. In this case, the coverage constraint of the EUA problem is ignored. In other words, every server can potentially serve all users. 

In the dataset, the first 15 cases are distributed and the last 15 cases are centralized. Note that the latter also have a larger problem size and on average greater DC ratio. The search space becomes larger, while the heuristics based on entity sizes apply better.

It is not so straightforward to evaluate a multi-objective optimization problem. Therefore, we take a weighted sum of the two objectives to be our final objective to minimize. 
\begin{equation}
\text{score} = - \frac{3}{n_u} \cdot A_u + \frac{1}{n_s} \cdot U_s
\end{equation}
This function has been used in previous works \cite{Falkestrom1770572},\cite{Allard1837710},\cite{Gustav} by Ericsson AB and in the Gurobi solutions in the dataset. We will refer to it as the score function or the score.

To benchmark performance, we compare our model’s score against that of the Gurobi Optimizer—a state-of-the-art solver for mathematical programming that leverages advanced algorithms and parallelism to find globally optimal solutions. The quality of our solution is measured by the percentage difference from the Gurobi (optimal) score, which we call the Performance Gap (PG). An average performance gap below 20\% is considered acceptable. Additionally, we analyze the distribution of user allocation and server usage to further assess the effectiveness of the proposed heuristic.

Since the dataset consists of two types of test cases, we also calculate the PG separately on them in order to better understand the performance of the model. 

\begin{figure*}[t]
    \centering
    \subfloat[]{%
    \includegraphics[width=0.49\textwidth]{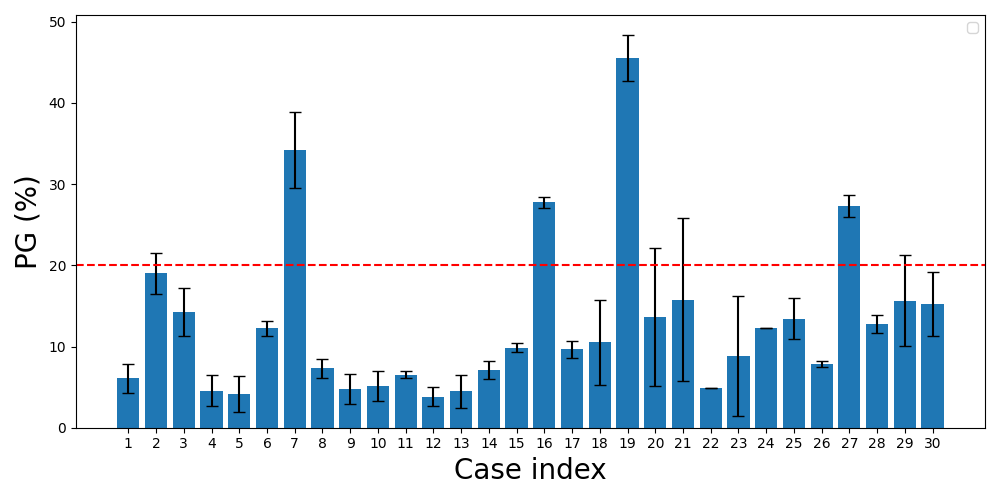}
    }\hfill
    \subfloat[]{%
    \includegraphics[width=0.49\textwidth]{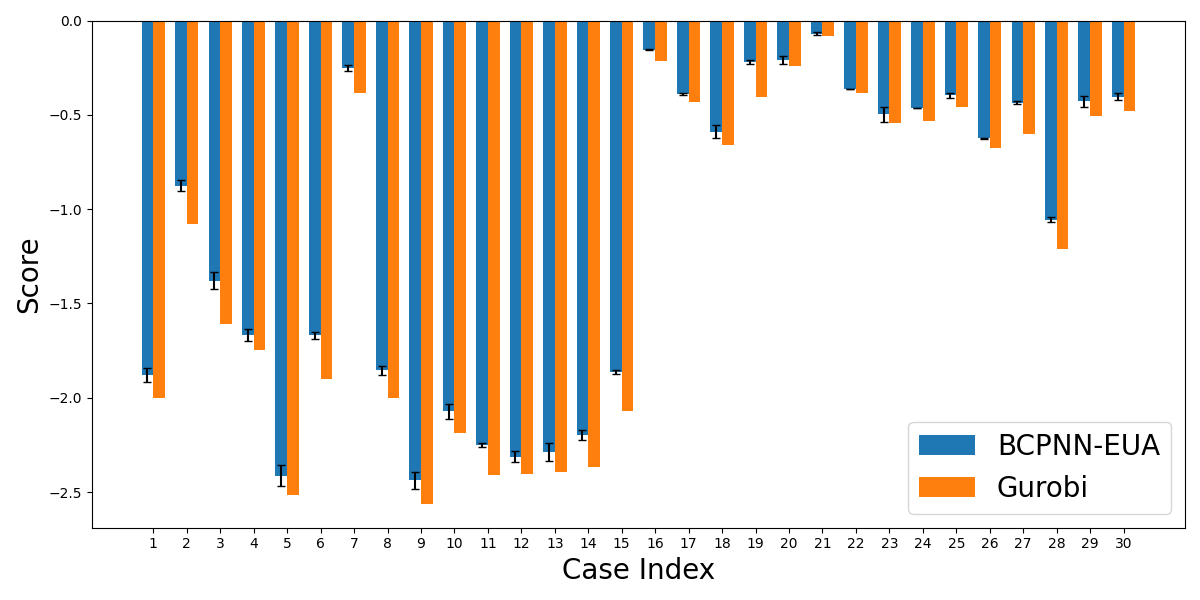}
    }
    \caption{Performance comparison between BCPNN-EUA and the optimal Gurobi solutions across all 30 test cases. 
(a) PG between BCPNN-EUA and Gurobi, with error bars indicating the standard deviation across five runs. The red dashed line marks the 20\% KPI threshold. 
(b) Raw scores of BCPNN-EUA and the corresponding Gurobi scores. The x-axis denotes the case index, and the y-axis the score for each case (lower is better, but scores are not comparable across different cases). Error bars on the BCPNN-EUA bars show the standard deviation across runs.
}
    \label{fig:performance}
\end{figure*}

\section{Results}
\subsection{Score and Performance Gap}
To evaluate whether the proposed BCPNN-EUA solver can effectively address the EUA problem, we ran the model 5 times for each test case and calculated the mean and the variance of the score of the solution from our model. Then the percentage difference between our model score and the score of the optimal solution was compared (Fig. \ref{fig:performance}).

Our model reaches an average of 12.8\% worse than the optimal solutions across the dataset, which is within the 20\% requirement. A Wilcoxon signed-rank test was conducted to compare the PG of the proposed method with the 20\% baseline. The results revealed a statistically significant difference with a large effect size (for distributed cases: W = 9.0, n = 15, p = 0.001, r = -0.75; for centralized cases: W = 31.0, n = 15, p = 0.054, r = -0.43), indicating that the proposed approach achieved significantly smaller PG overall. The model achieves lower PGs at the distributed, low DC ratio cases (9.56\% PG) than the centralized, high DC ratio cases (16.07\%). We notice that the variance of PG is high for some cases (e.g. 20, 21), likely due to them having a Gurobi score close to zero and thus the PG is more sensitive to changes in the allocation plan. For most instances, the scores produced by BCPNN-EUA closely track the optimal Gurobi scores (r = 0.9980, p = 5.1176e-35, n = 30 for the Pearson correlation test) , and the variability across runs is small. This indicates that, despite the stochastic WTA dynamics, the solver is numerically stable and converges to solutions with acceptable qualities.

\subsection{Timestep to Solution}
Besides solution quality, it is also important to assess how much time the model requires to generate a solution. Owing to limited access to neuromorphic hardware, we conducted all experiments solely via software simulations and used the number of simulation time steps as a proxy for time.

Due to the stochastic nature of the update process and the design of the loadbias function, the activation may flip between some states, normally when some servers only have a small remaining space left, but allocating another user to it will cause overfilling. Those flipping states will result in the same score after postprocessing; therefore, we examined the score of the solution to estimate the number of time steps needed.

\begin{figure}[t]
    \centering
    \begin{minipage}[b]{0.48\linewidth}
        \centering
        \includegraphics[width=\linewidth]{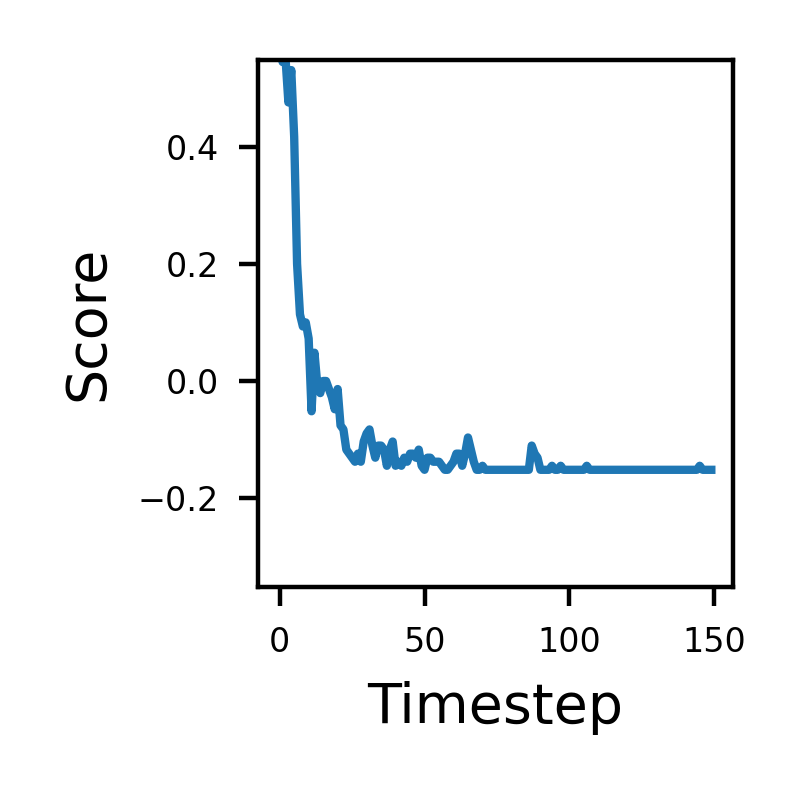}
    \end{minipage}
    \begin{minipage}[b]{0.48\linewidth}
        \centering
        \includegraphics[width=\linewidth]{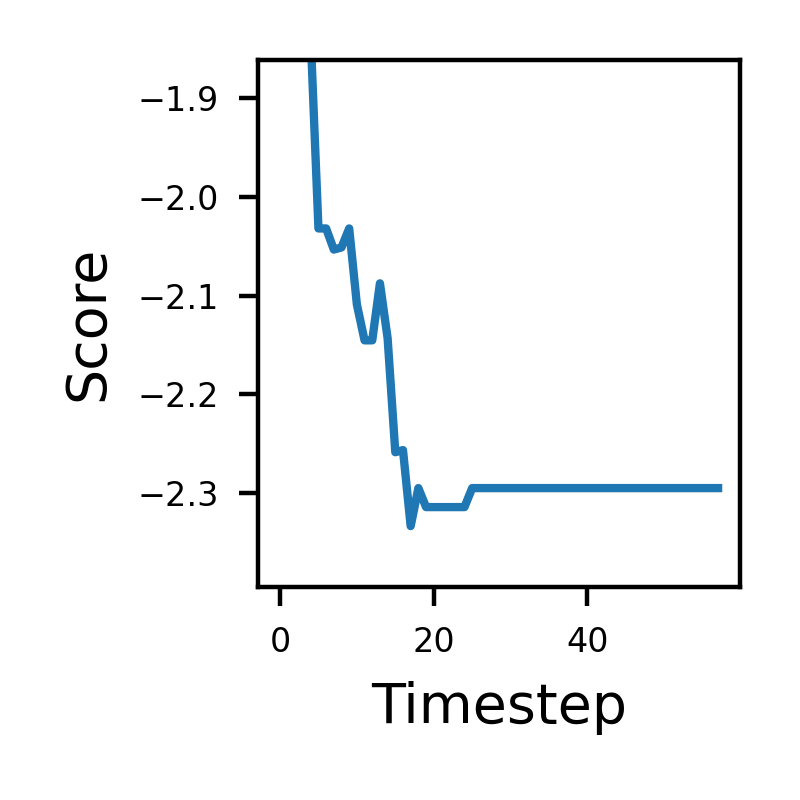}
    \end{minipage}
    \caption{Score evolution over timesteps for two representative runs of the BCPNN–EUA model.}
    \label{timesteps}
\end{figure}

Judging from experimental results, it would take at maximum around 150 timesteps for each turn with a fixed $b_0$ to converge, some example plots are shown in Fig. \ref{timesteps}. The test results used 6 different $b_0$ values: -180, -50, -40, -30, -20, -10, so the model will take at most 900 timesteps to generate a solution.



The network dynamics evolve rapidly in the early steps and gradually stabilize over time. This behavior is driven by the dominance of the static component of the external heuristic input, which outweighs the stochastic and dynamic influences during updates.

\subsection{Effect of Heuristic Components and Parameter Sensitivity}
\begin{figure*}[t]
    \centering
    \subfloat[]{%
     \includegraphics[width=0.32\textwidth]{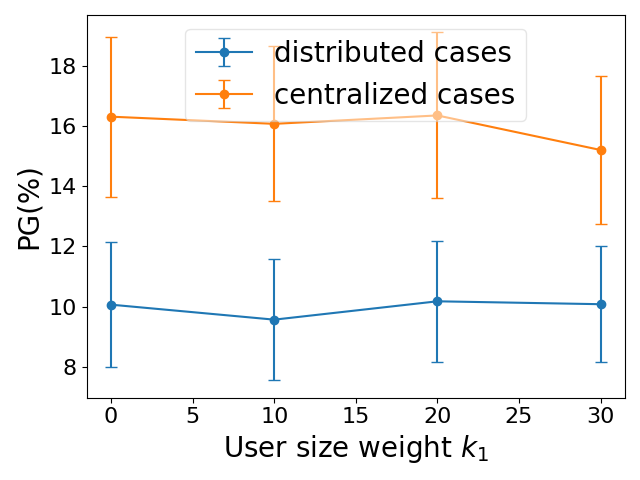}
    }\hfill
    \subfloat[]{%
    \includegraphics[width=0.32\textwidth]{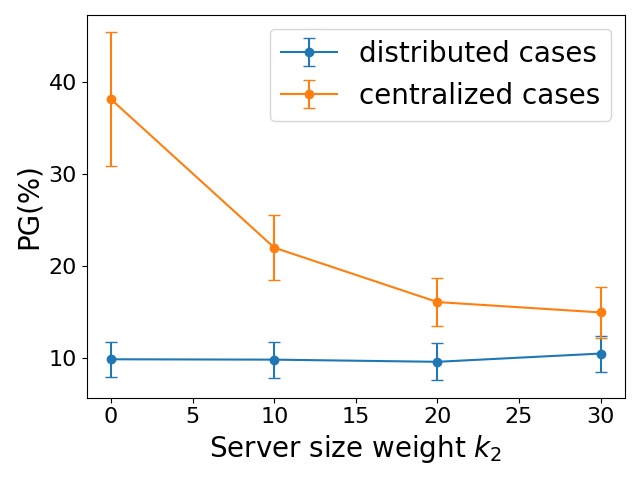}
    }\hfill
    \subfloat[]{%
    \includegraphics[width=0.32\textwidth]{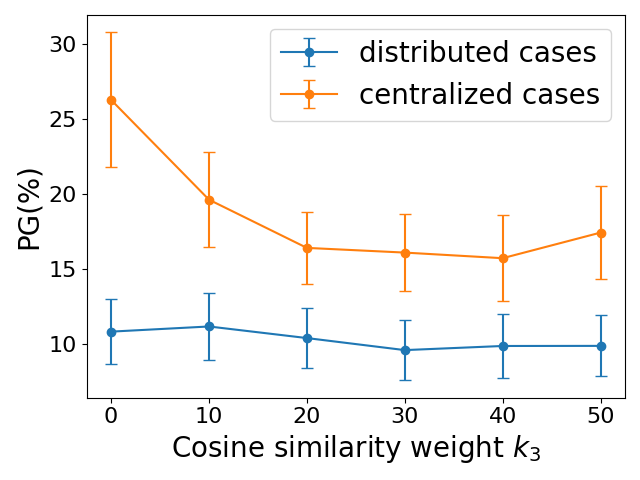}
    }
    \caption{Parameter sensitivity tests for the heuristic weights. The panels show the average performance gap (PG) for different values of (a) the user size weight $k_1$, (b) server size weight $k_2$, and (c) cosine similarity weight $k_3$, respectively, for distributed and centralized cases, with the standard error as the error bar. The reference configuration is given by $k_1 = 10$, $k_2 = 20$, and $k_3 = 30$, and all other points are obtained by varying a single parameter around this reference while keeping the remaining two fixed.
}
    \label{fig:heuristic_weights}
\end{figure*}

To examine the effectiveness of the heuristic components, we varied the coefficients of the terms in (20). To understand how each individual component contributes to the overall performance of the model, we systematically modified one weight parameter at a time, while keeping the rest of the model unchanged. Each modified configuration was evaluated on the same set of test cases as the full model, and the PG was recorded separately for distributed and centralized cases. The results of these experiments (Fig. \ref{fig:heuristic_weights}) indicate which mechanisms are most critical for the model's ability to find high-quality solutions.

Overall, we did not observe large performance changes under small perturbations of the parameters, which suggests that the model is reasonably robust. For the distributed cases in particular, the PG remains almost unchanged across different parameter settings.

Somewhat surprisingly, the weight on the user-size term has only a minor effect on the PG, even though the final allocation plans clearly tend to favor users with smaller demands. This can be partly explained by properties of the dataset and our definition of “size”. In the distributed cases, the DC ratios are mostly below one, so the system does not need to drop any users and the exact ordering among them matters less. In the centralized cases, both CORE and RAM DC ratios are high, and we define the user size using only the scarcer resource. In such situations, servers may become saturated along the other resource dimension first, so preferring users with a low demand in the selected dimension does not necessarily translate into serving more users overall.

By contrast, large degradations occur on the centralized cases when either the server-size component or the cosine-similarity term is removed. This matches our intuition: when the aggregate user demand exceeds the total server capacity, prioritizing larger capacity servers helps increase the number of served users, and taking both resource constraints into account allows the model to utilize server capacity more fully. On the other hand, setting these weights too high also harms performance, likely because they distort the loadbias curve, which is the key mechanism that encourages servers to be filled close to, but not substantially beyond, their capacity.

Across all three panels, the error bars remain relatively small and change only mildly as the parameters are varied. For the distributed cases, the spread is low and almost constant, indicating that the solver behaves consistently across parameter settings on the easier instances. For the centralized cases, the variability is larger, as expected for harder problems, but it does not fluctuate strongly except at extreme values where a component is effectively removed (e.g., $k_2=0$ or $k_3=0$), which also coincides with higher mean PG. Taken together, this suggests that the method is reasonably robust to moderate changes of the heuristic weights.

\subsection{Computational and Energy Considerations}

In terms of computational demand, the heuristic components add only a modest overhead on top of the core BCPNN dynamics. The loadbias and the cosine similarity term requires reading out the current assignment to compute server utilization, which scales as $O(n_u n_s)$. The user-size and server-size terms are static features that are precomputed once per instance and then applied as simple scalar factors when generating the external bias, without changing the overall complexity. The cosine-similarity term involves computing the residual capacity vector of each server and a two-dimensional dot product for each user–server pair, which also scales linearly in $n_u n_s$ with a constant factor. Lateral inhibition is implemented as fixed local recurrent connections within each WTA motif and does not introduce any additional global optimization loop. As a result, the added heuristics improve solution quality without incurring a prohibitive computational cost.





\begin{figure}[htbp]
\centerline{\includegraphics[width=\linewidth]{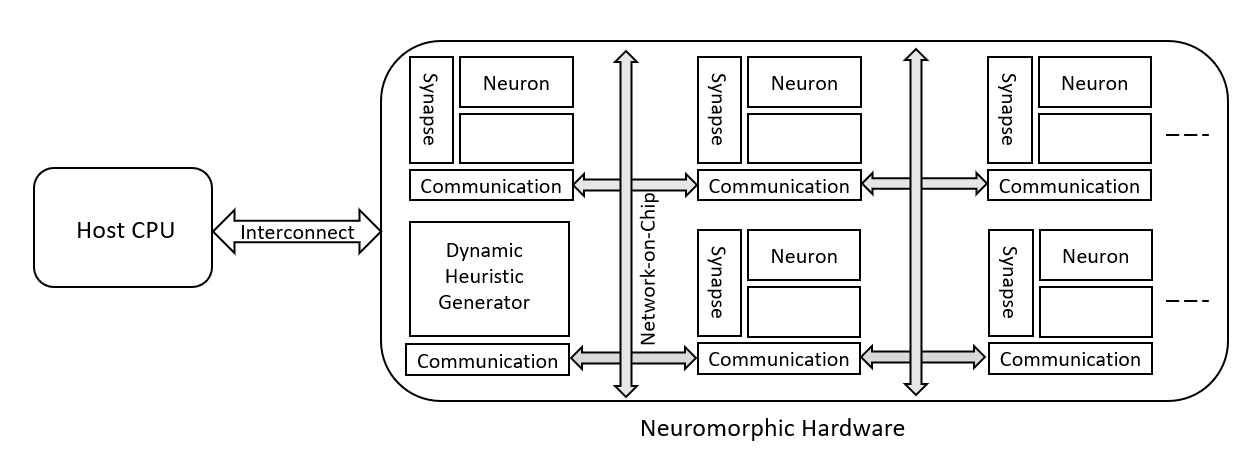}}
\caption{Conceptual Figure for Neuromorphic Implementation}

\label{fig:neur_HW}
\end{figure}

We propose implementing the WTA motifs on neuromorphic hardware composed of programmable neuron cores with hardware support for exponential operations.
A dedicated on-chip configurable logic unit (CLU) is employed to execute the dynamic heuristic generator, while a high-speed network-on-chip enables communication between neuron cores and specialized units, supporting the exchange of graded spike signals. This architecture (see an example in Figure~\ref{fig:neur_HW}) is consistent with existing platforms such as Loihi 2, where embedded processors interface with neuron cores via the same on-chip network. While embedded processors offer flexibility and programmability for implementing heuristic components, they introduce trade-offs in terms of parallelism, latency, and energy efficiency. In contrast, a dedicated CLU can exploit parallel and pipelined computation across user–server pairs to accelerate heuristic execution. This unit operates in close synchronization with the WTA dynamics of the neuromorphic cores, minimizing latency between consecutive simulation time steps. We further note that the proposed solution can be adapted to spike-based WTA motifs consisting of excitatory and inhibitory neurons, which have been successfully mapped to Loihi~\cite{uludaug2023exploring}. 

The host is responsible for compiling the byte-stream for each mapped neuron and synapse, as well as for the dynamic heuristic generator. Using the runtime, the host configures these byte-streams onto the respective neuromorphic cores and the CLU. It also streams the problem-instance-specific context—including user demands, server capacities, and uncovered users for each server—to the CLU and reads out the final allocation after a predetermined number of simulation time steps. This approach enables the dynamic heuristic generator to handle changes in UE locations, fluctuations in individual UE resource demands, and variations in available edge server capacities, while keeping the total number of users and servers fixed due to the compilation and mapping overhead of the WTA motifs on the neuromorphic cores.
.

We estimated the average energy consumption across the problem instances when solved on the multi-core CPU and on neuromorphic hardware like Loihi2. The TDP of CPU used is 15W\cite{intel_i5_1245u} , thus the energy consumption is 15 * 43.84 = 657.7J. The neuromorphic sources (neurons, synapses) require to map WTA circuits of the problem instances can be met with a Loihi2 chip (1M neurons, 120 million synapses), which has a TDP of 2.26W (The Hala Point system — which integrates 1,152 Loihi2 chips — consumes a maximum of 2,600W for the full system\cite{intel_hala_point}. The lower bound for inter-spike arrival time on the same hardware is 200ns. As the proposed method takes 150 time ticks on average to solve the problem instances, the energy consumption in the best case could be 2.26 * 150 *200ns = 67.8uJ. We note that this estimate excludes the energy overhead of dynamically generating heuristics, as that process may require specialized support from neuromorphic hardware.

\label{tab:ablation}

\section{Discussions}

The model decouples the allocation objective into two parts: filling a fixed set of servers efficiently, and tuning the number of active servers through a global parameter $b_0$. This dual-level decomposition supports flexible tradeoff control between user inclusion and server usage, without embedding all constraints into a static energy function.

A real-life application setting could potentially be dynamic, for example, users may move, join, or leave the field. While our current implementation assumes static input, a promising extension would involve starting each simulation from the previous solution and updating allocations incrementally. This could further reduce convergence time, especially when user demands or server states shift slightly. Although we lack real-world time-series EUA datasets to evaluate this, the model structure is well suited for such settings, as stochastic attractor dynamics naturally support continual evolution.

Another consideration regarding hardware implementation is the problem size, since the network size varies across cases. One simple solution would be to instantiate a network with the maximum required dimensions and mute the unused units via strong negative input, similar to how uncovered units are handled. This allows for reuse of the same neuromorphic hardware across problem instances without reconfiguration.

Experimental results show that the DC ratio—especially when both resources are tight—has a stronger influence on performance than problem size. In high DC-ratio centralized cases, the cosine similarity term becomes critical, aligning allocation choices with residual capacity. Its removal causes significant degradation in such instances, affirming the need for dual-resource alignment.

Although attractor networks traditionally rely on energy descent dynamics\cite{hopfield},\cite{hopfield2},\cite{ISHII1997941}, modeling the full EUA objective within an energy function introduces nontrivial consistency constraints that are hard to encode in the weight matrix. More specifically, the server usage status depends on the joint activity of user allocation units, resulting in high-order terms that cannot be captured via pairwise weights alone. Existing attempts to model this can not avoid opening unused servers unless post-processing is applied. Moreover, the loadbias component of our dynamic inputs, which is crucial for real-time server loading control, breaks the assumption of a stable energy landscape\cite{BCPNNenergy}. Therefore, we adopt a heuristic-driven method, scanning over $b_0$ to simulate different user-server tradeoffs, effectively replacing energy descent with a parameter sweep.


The model is well suited for real-time and energy-constrained deployment. Its weight matrix is sparse, input size scales with the number of units, and local update dynamics minimize communication and computational cost. The modular BCPNN structure maps naturally to neuromorphic hardware and has been implemented on FPGAs \cite{FPGA} and memristor arrays \cite{BCPNNmemristor}. Unlike many prior models, our design scales to larger problem instances without parameter tuning.


Nonetheless, the current heuristics are manually crafted and may underperform under distribution shift. Some components, such as loadbias, involve nonlinear computations that could be hardware-expensive, suggesting future work on approximation methods.


\newpage
\vspace{12pt}

\bibliographystyle{IEEEtran}
\bibliography{references}
\end{document}